\documentclass[sigconf, 10pt, letterpaper]{acmart}

\usepackage{amsmath}
\usepackage[ruled]{algorithm2e} 
\usepackage{balance}
\usepackage{subfig}
\usepackage{wrapfig}
\usepackage{mathtools}
\usepackage{graphicx}  

\usepackage{caption}
\renewcommand\footnotetextcopyrightpermission[1]{} 
\setcopyright{none}
\begin{document}
\title{FedPrune: Towards Inclusive Federated Learning}

\author{Muhammad Tahir Munir, Muhammad Mustansar Saeed, Mahad Ali, Zafar Ayyub Qazi, Ihsan Ayyub Qazi}
\affiliation{
	\institution{Department of Computer Science, LUMS}
}
\email{{18030016, 18030047, 21100119, zafar.qazi, ihsan.qazi}@lums.edu.pk}

\begin{abstract}
Federated learning (FL) is a distributed learning technique that trains a shared model over distributed data in a privacy-preserving manner.
Unfortunately, FL's performance degrades when there is (i) variability in client characteristics in terms of computational and memory resources (system heterogeneity) and (ii) non-IID data distribution across clients (statistical heterogeneity). For example, slow clients get dropped in FL schemes, such as Federated Averaging (FedAvg), which not only limits overall learning but also biases results towards fast clients.
We propose \textsf{FedPrune}; a system that tackles this challenge by pruning the global model for slow clients based on their device characteristics.
By doing so, slow clients can train a small model quickly and participate in FL which increases test accuracy as well as fairness.
By using insights form Central Limit Theorem, \textsf{FedPrune} incorporates a new aggregation technique which achieves robust performance over non-IID data. Experimental evaluation shows that \textsf{FedPrune} provides robust convergence and better fairness compared to Federated Averaging.

\end{abstract}

\maketitle
\section{introduction}
Today's deep neural networks (DNNs) power a wide variety of applications ranging from image classification, speech recognition, to fraud detection \cite{dnn}.
DNNs rely on large amounts of data to make accurate predictions and draw useful inferences.
However, such data is often \emph{private}\footnote{Such data is also known as Personally Identifiable Information (PII).} and thus not readily available due to rising privacy concerns and emerging privacy regulations (e.g., Europe's GDPR \cite{gdpr} and California Consumer Privacy Act \cite{ccpa}). Lack of useful data can severely limit the effectiveness of DNNs.

Federated learning (FL) is a distributed machine learning technique that tackles this problem by training a global model over data distributed across multiple edge devices (e.g., mobile phones) and sharing the model parameters with a centralized server to aide learning in a privacy preserving manner \cite{fl,guideFL}.
Unfortunately, FL's effectiveness degrades when model training involves heterogeneous client devices \cite{openfl,flsys}; a common case especially in developing countries \cite{imc_sohaib,mobile_ccr}.
Slow clients are dropped in FL from the training process, which not only limits overall learning for everyone (e.g., by degrading test accuracy and reducing fairness \cite{agnosticFL}) but also biases results and degrades user experience due to systematic exclusion of slow clients.

We propose \textsf{FedPrune}; a system that tackles this challenge by \emph{adapting} the model size based on client device capabilities.
Thus, slow clients are served smaller models whereas faster clients train on larger models.
Such an approach offers two key benefits: (i) For IID data, faster clients can enable learning for slow clients including in parts of the model architecture where the slow clients did not train over. This is unlike approaches (e.g., Federated Dropout \cite{flclient}), in which all clients are served a subset model of the \emph{same} size and (ii) for non-IID data, serving small models to slow clients improves task accuracy as well as fairness while being inclusive (i.e., without dropping of clients).

To improve robustness over non-IID data (statistical heterogeneity), \textsf{FedPrune} calculates model parameters for every round of FL using insights from Lyapunov's Central Limit Theorem (CLT) \cite{feller}, which posits that the distribution of the sample mean of independent random variables (which need not necessarily be from \emph{same} distribution) converges to a Normal distribution.

To inform data-driven sub-model selection, \textsf{FedPrune} initially picks a \emph{random} sub-model and asks each slow client to train over the model. After $r$ rounds, \textsf{FedPrune} uses the average post-activation values to sort neurons in each of the hidden layers of the centralized model and then picks a new sub-model based on post activation values.\footnote{In case of CNN models, we also drop filters.}
This allows neurons with little or no activations to be excluded from the sub-model.

We carry out extensive evaluation using (i) large-scale simulations involving the LEAF benchmarking framework for learning in federated settings \cite{leaf} and (ii) small-scale real testbed experiments involving mobile clients with heterogeneous device capabilities.
Experiments show that \textsf{FedPrune} achieves better test accuracy than Federated Averaging (FedAvg) \cite{fl} across several datasets and models including feed-forward neural networks (FNNs), convolution neural networks (CNNs) as well as recurrent neural networks (RNNs). In particular, in settings with systems and data heterogeneity, \textsf{FedPrune} shows significant advantages in terms of convergence and test accuracy relative to FedAvg.
\begin{figure}
    \centering
    \includegraphics[width=0.75\linewidth,scale=0.35]{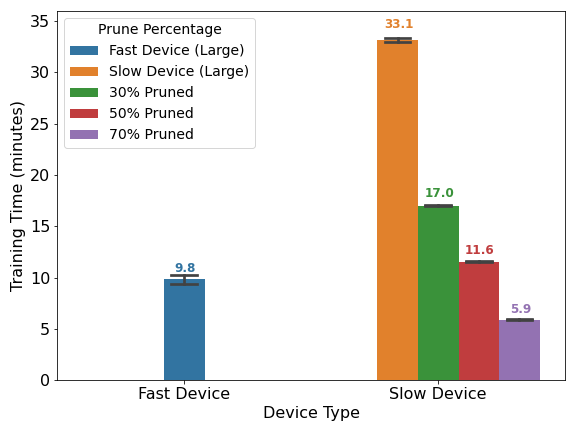}
    \vspace{-0.15in}
    \caption{Model training times on slow and fast clients. Slow device is the Nokia 1 smartphone (Quadcore, 1\,GB RAM) and the fast device is Nexus 6P (Octacore, 3\,GB RAM).}
    \label{fig:time_between_slow_fast}
    \vspace{-0.15in}
\end{figure}

Taken together, we make the following contributions in this work:
\begin{itemize}
    \item We design \textsf{FedPrune}; an adaptive model serving framework for FL that adapts model sizes based on client capabilities (\S \ref{sec:design}).
    \item To achieve better generalization, we propose a CLT-based approach, which outperforms other approaches including FedAvg (\S \ref{subsec:clt}).
    \item We carry our extensive evaluation using large-scale simulations as well as small-scale real testbed experiments over a wide variety of real-world federated datasets. Our results show that \textsf{FedPrune} achieves robust performance across a range of scenarios (\S \ref{sec:eval}).
\end{itemize}

\section{Problem Motivation}
It is common for distributed clients in FL to exhibit considerable heterogeneity in terms of computational resources (e.g., CPU cores), network bandwidth, and energy \cite{flclient,fedprox}. This heterogeneity impacts the accuracy and training time of the FL process, which commonly involves training complex deep learning models comprising large number of parameters.
Consequently, resource-constrained devices (e.g., entry-level smartphones \cite{mobile_ccr,theo_memory}) are either unable to train the models due to limited computational and memory resources or take a prohibitively long time in training. Such slow clients (or stragglers) are dropped for efficiency reasons in traditional FL, which can (i) degrade test accuracy and (ii) lead to unfairness. On the other hand, waiting for slow clients to report their updates can increase convergence times and degrade user experience.
\begin{figure*}
  \includegraphics[width=1.0\linewidth,scale=0.5]{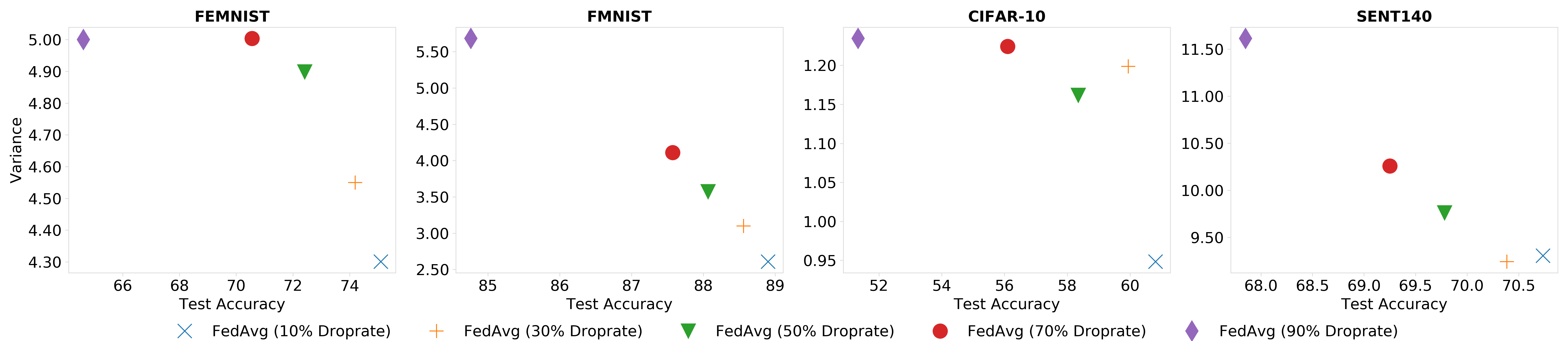}
  \vspace{-0.2in}
  \caption{Impact of systems heterogeneity on the robustness and fairness in FedAvg. As the number of slow clients starts to increase, model performance degrades.}
  \label{fig:droprates_variance_test_accuracy_all}
  \vspace{-0.1in}
\end{figure*}

To quantify the impact of serving a large model to slow clients, we train a CNN model (for details, see Section \ref{leaf_model_architecture}) for the FEMNIST dataset \cite{leaf}, and a simple feed-forward neural network for a synthetic dataset (see Appendix \ref{synthetic_results}), and measure the training time on real mobile devices, i.e., Nokia 1 (Quadcore, 1\,GB RAM) and Nexus 6P (Octacore, 3\,GB RAM). These devices represent slow and fast clients, respectively, and are popular in developing countries \cite{mobile_ccr}.

We quantify the training time in FL by varying the model size over which the slow client trains while always serving the large model to the fast client.
Figure \ref{fig:time_between_slow_fast} shows the average training time (per round) for both the devices.
The results show that the slow device takes 3.4$\times$ longer time compared to the fast device when training the same large model.
When served with the 50\% pruned model, the slow device saw a $\sim$2.9$\times$ reduction in training time (i.e, training time reduced from 33.1\,mins to 11.6\,mins).
Observe that this improved training time (using 50\% model drop rate) is within $\sim$1.1$\times$ of fast device's training time.
We also observe similar improvements in training times on a \textit{synthetic} dataset with a 1 million parameter FNN. In particular, the slow device takes $\sim$1.5$\times$ more time compared to the fast device. We also observe a $\sim$2$\times$ reduction in training times when using a 50\% pruned model (see Appendix \ref{synthetic_results}).
\begin{figure}[t]
    \centering
    \includegraphics[width=0.75\linewidth,scale=0.35]{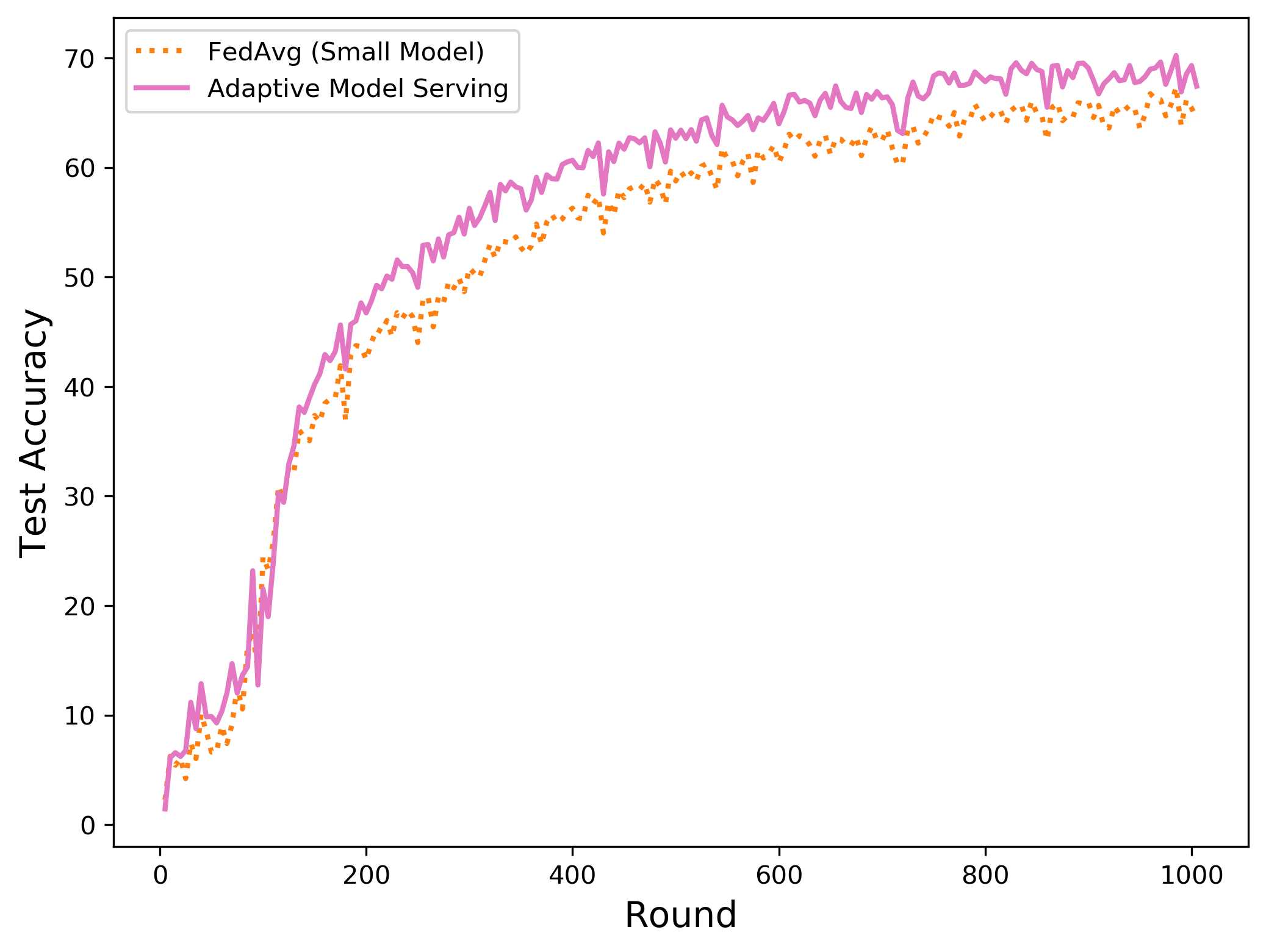}
    \vspace{-0.1in}
    \caption{Small vs. Adaptive Model Serving.}
    \label{fig:adaptive_model_vs_fedavg_small}
    \vspace{-0.1in}
\end{figure}

\begin{figure*}[t]
  \includegraphics[width=1\linewidth,scale=0.5]{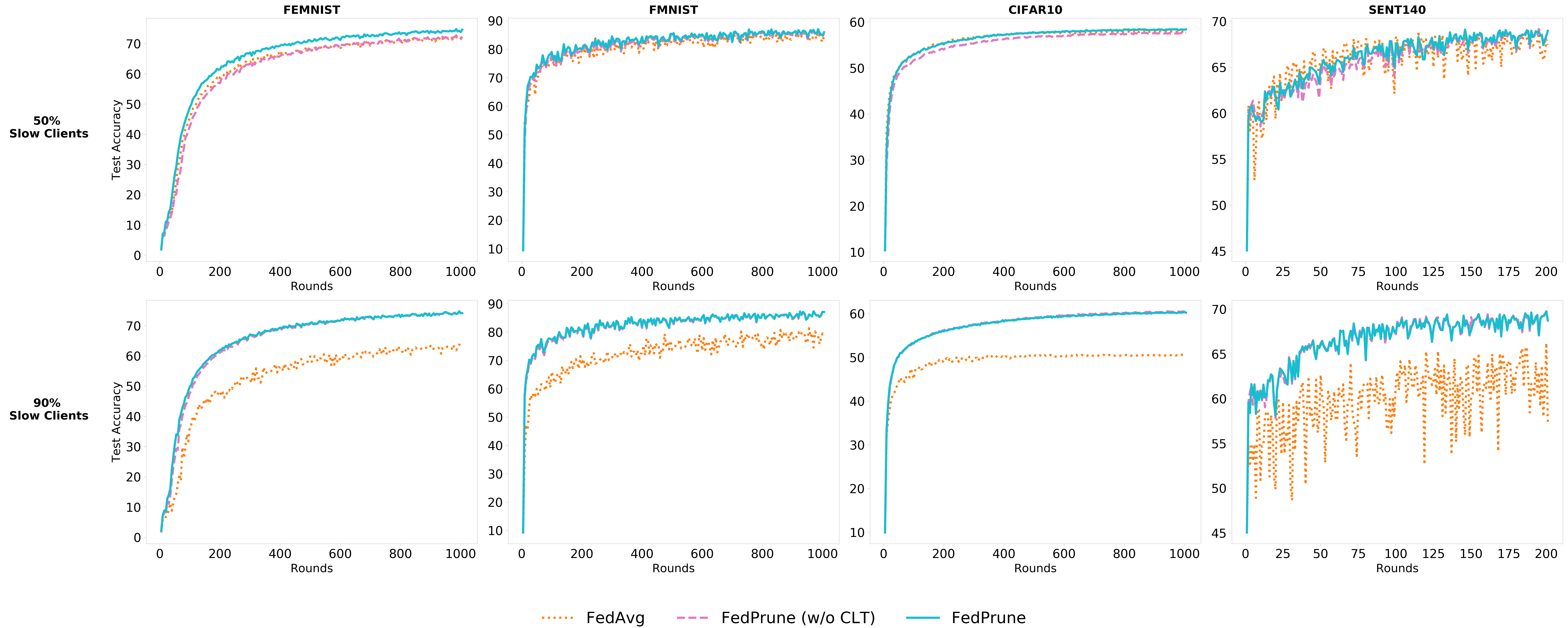}
  \vspace{-0.2in}
  \caption{\textsf{FedPrune} results in significant convergence improvements relative to FedAvg in heterogeneous networks. We simulate different levels of systems heterogeneity by forcing 50\%, and 90\% of devices to be the slow devices (dropped by FedAvg). Comparing FedAvg and \textsf{FedPrune}, we see that serving a smaller model to slow clients can help convergence in the presence of systems heterogeneity. We also report train loss in Figure \ref{fig:train_loss_all}, Appendix \ref{simulated_experiments} }
  \label{fig:test_accuracy_all}
  \vspace{-0.1in}
\end{figure*}

We observe that the slow device takes a prohibitively long time (i.e., 33.1\,mins) in our testbed, and if we wait for such a client then the overall training will be bounded by this client. However, if we drop such slow clients for efficiency purposes then the test accuracy is impacted as shown in Figure \ref{fig:test_accuracy_all}. \textsf{FedPrune} provides an opportunity by serving the small model to slow clients so that they can train and report their updates within a time budget, and our results show that we obtain significant improvement in model training time by serving the small model to the slow client.

To further evaluate the impact of slow clients, we simulate different levels of system heterogeneity in LEAF. In particular, we emulate 10\%, 30\%, 50\%, 70\% and 90\% slow clients and train them on multiple real datasets, i.e., FEMNIST, FMNIST, CIFAR-10, and Sent140 (see Section \ref{sec:eval} for details). Figure \ref{fig:droprates_variance_test_accuracy_all} shows that in FedAvg, system heterogeneity negatively impacts the robustness and fairness of the model. In FedAvg, as the number of slow devices increases, the performance starts to degrade.
Thus, serving a single model (of the same size) presents a tradeoff.
If we serve a large model to all clients, slow clients may not be able to train the model, which will degrade model's performance (see Figure \ref{fig:droprates_variance_test_accuracy_all}).
However, if we serve a small subset model to all clients, due to complexity of datasets, it may perform poorly. For empirical evidence, we train a subset model (50\% pruned) on all clients using the FEMNIST dataset and compare it with \textsf{FedPrune} w/o CLT, which serves a sub-model to slow clients only. 
Figure \ref{fig:adaptive_model_vs_fedavg_small} shows that adaptively serving subset model to only slow clients performs better than serving the small model to all clients.

\section{Background and Related Work}
In this work, we focus on synchronous federated learning (FL) algorithms that proceed in rounds of training. These algorithms aim to learn a shared global model with parameters embodied in a real tensor $\boldsymbol\Gamma$ from data stored across a number of distributed clients.
In each round $t \geq 0$, the server distributes the current global model $\boldsymbol\Gamma_t$ to the set of selected clients $S_t$ with a total of $n_t$ data instances.
The selected clients locally execute SGD on their data and independently update the model to produce the updated models \{$\boldsymbol\Gamma_t^k  | k \in S_t$\}. The update of each client $k$ can be expressed as:
\begin{equation}
    \boldsymbol\Gamma_t^k = \boldsymbol\Gamma_t - \alpha H_t^k, \,\,\, \forall k \in S_t 
\end{equation}
where $H^k_t$ is the gradients tensor for client $k$ in training round $t$ and $\alpha$ is the learning rate chosen by the server. Each selected client $k$ then sends the update back to the server, where the new global model ($\boldsymbol\Gamma_{t+1}$) is constructed by aggregating all client-side updates as follows:
\begin{equation}
    \boldsymbol\Gamma_{t+1} = \sum_{\forall k \in S_t} \frac{n_k}{n_t} \cdot \boldsymbol\Gamma_t^k
\end{equation}
where $n_k$ is the number of data instances of client $c$ and $n_t = \sum_{\forall k\in S_t} n_k$. Hence, $\boldsymbol\Gamma_{t+1}$ can be written as:
\begin{equation}
    \boldsymbol\Gamma_{t+1} = \boldsymbol\Gamma_t - \alpha_t H_t
\end{equation}
where $H_t = \frac{1}{n_t} \sum_{k \in S_t} n_k H^k_t$.

Our goal in this paper is to come up with a framework for reducing the computational and memory overhead of models served to slow clients.
Our work is motivated by the observation that increasingly clients, especially in developing countries, are constrained by their hardware resources (e.g., memory, number of CPU cores) \cite{mobile_ccr,imc_sohaib,imc16_video,android_memory_google}.

\textbf{Fairness in FL}. 
Due to the heterogeneity in client devices and data in federated networks, it is possible that the performance of a model will vary significantly across the network. This concern, also known as representation disparity, is a major challenge in FL, as it can potentially result in uneven outcomes for the devices.
Following Li et al. \cite{ditto}, we provide a formal definition of this fairness in the context of FL below.

\textbf{Definition.} \emph{We say that a model $W_1$ is more fair than $W_2$ if the test performance distribution of $W_1$ across the network is more uniform than that of $W_2$, i.e., $std\{F_k(W_1)\}_{k\in[K]} < std\{F_k(W_2)\}_{k\in[K]}$ where $F_k(\cdot)$ denotes the test loss on device $k\in[K]$, and $std\{\cdot\}$ denotes the standard deviation.}

We note that there exists a tension between variance and utility in the definition above; in general, the goal is to \emph{lower} the variance while maintaining a reasonable average performance (e.g., average test accuracy).
Several prior works have separately considered either fairness or robustness in federated learning. For instance, fairness strategies include using minimax optimization to focus on the worst-performing devices \cite{asyncFL,multi-fairness} or reweighting the devices to allow for a flexible fairness/accuracy tradeoff (e.g., \cite{fairq}). 

\textbf{Reducing model size}. To reduce the neural network complexity, different techniques have been proposed in the literature. Diao et al. \cite{diao2020heterofl} proposed a framework HeteroFL to reduce the computation and communication cost by training local models created from a single global model based on clients' device characteristics. Jiang et al. \cite{jiang2019model} proposed the adaptive and distributed model pruning technique (PruneFL) in which it performs the initial pruning by picking a client randomly using its local data, and further adaptive pruning using importance measure to adapt the model size during the training process. The limitation of this work includes: (i) initial pruning can be biased towards the selected client (ii) initial pruned model may not load into the memory if there are more resource constrained devices (e.g., less RAM, CPU cores) in the pool than the selected one. Tan et al. \cite{tan2020dropnet} proposed an iterative approach (DropNet) in non-FL settings to remove neurons and filters as the training progresses.
\section{design}
\label{sec:design}
\textsf{FedPrune} has the following key design features:

\begin{enumerate}
    \item \emph{Differential model serving}. Clients are served models based on their capabilities. Thus, slow clients are served a small model whereas fast clients are served large models.
    \item \emph{Sub-model selection using activations.} Small sub-models are selected for slow clients based on post-activation values.
    \item \emph{Model generalization using insights from the Central Limit Theorem}. To improve performance robustness especially over non-IID datasets, we use a CLT based approach to choose model parameters.
\end{enumerate}

\subsection{Differential Model Serving (DMS)}
DMS has several benefits compared to serving the \emph{same} model to all clients participating in FL.
First, due to the large heterogeneity in mobile device characteristics, training a single model over all clients can lead to widely different training times, which can result in frequent dropping of slow clients from FL\footnote{Note that some clients may not be able to run large models at all due to memory constraints.}, potentially leading to unfairness across clients \cite{flclient,flsys,ditto}.
Second, it is difficult to choose a single model that allows all clients to participate in FL training while achieving high accuracy; small models can degrade model accuracy whereas large models lead to dropping of slow clients.

DMS addresses these challenges by allowing model sizes to be adapted based on device characteristics such as number of CPU cores, memory size, and GPU characteristics (if present). Thus, slow clients are served smaller models than faster clients. This improves fairness by reducing dropping of clients from the FL process.
In case of IID datasets, large clients can help slow clients by sharing model weights not included in the sub-model served to slow clients.
When the dataset is non-IID, DMS incorporates learning of all clients unlike schemes that drop clients altogether.

Existing schemes like FedProx \cite{fedprox} serve the same global model to all clients and incorporate partial work done by slow clients. It assumes that slow clients can train for at least a few epochs. In reality, some clients may not be able to train a large model at all. Moreover, incorporating partial work based on only a few epochs can slow convergence (e.g., training loss may be too high when trained for only a few epochs).
Schemes like Federated Dropout \cite{flclient} serve models of the same size to all clients in each round. Again, the subset model may be too large for some slow clients to complete training, which can lead to dropouts. Moreover, it is unclear if we can pick one-size-fits-all model, which simultaneously achieves high accuracy and reasonable model training time.

\subsection{Sub-model Selection}
Given a model size (e.g., defined in terms of the dropout rate), a key design question in \textsf{FedPrune} is, ``\emph{which sub-model should we serve to slow clients?}".
Clearly, there are many possible sub-models one can pick.
In a feed-forward network, suppose we allow dropping of neurons from all layers including the input and output layers, then the number of distinct models are lower bounded by $\binom{n_1}{\lfloor f.n_1 \rfloor}\binom{n_2}{\lfloor f.n_2 \rfloor}..\binom{n_l}{\lfloor f.n_l \rfloor}$.
Thus, it is challenging to find the best model architecture among such a large set of possible sub-models.

We address this challenge by two strategies.
First, we bootstrap the FL process by choosing a random sub-model and allowing slow clients to train over the model for the first $r$ rounds.
This allows us to assess the parts of sub-model that contribute the most to the machine learning task.
Second, after $r$ rounds we choose a new subset model based on the post-activation values of neurons (in case of CNN models, filters too) \cite{dropnet}.
In particular, we sort neurons based on their post activation values and pick the top $m$ neurons that satisfy the model size constraint.
This allows neurons with no or small activations to be excluded from the subset model as well as allows neurons with large activations trained by only fast clients to be included in the subset model.
Thus, this approach enables better model selection than randomly picking models in each round or choosing a different random model in each round.\cite{flclient}

When combing activations from slow and fast clients, it is important to combine them carefully.
In particular, model parameters that are trained by both slow and fast clients will have more samples to be averaged over compared to model parameters involving only fast clients.
Thus, fast clients may dominate in average post-activation values or significantly impact sub-model selection (e.g., if their neurons end up high in the sorted order of post activation values).
To address this issue, we assign equal weight to model parameters trained by slow clients and fast clients.
However, these weights can be adapted to prioritize relative learning by slow and fast clients (e.g., based on whether the dataset is IID or non-IID).

\subsection{Model Generalization}
\label{subsec:clt}
Training a model in FL is challenging due to the statistical variations in data samples distributed across clients, which impacts both model accuracy as well as model convergence \cite{fedprox,flclient}.
This is a particularly challenging problem when the data is non-IID and
is exacerbated by the fact that in each round, FL picks $k$ \emph{random} clients for training from a pool of $N$ clients.
To generalize model training, we appeal to the Lyapunov's Central Limit Theorem (CLT), which posits that the distribution of the sample mean of independent random variables (which need not necessarily be from \emph{same} distribution) converges to a Normal distribution.

We consider the setting, where each client $i$ draws independent samples from a distribution $D_i$ with finite mean $\mu_i$ and finite variance $\sigma_i$. Let $X^i_j \sim D_i$ be the random variable denoting weight of the $jth$ model parameter for client $i$. Then FL aims to learn the average $X_j = \sum_{i=1}^{N} p_iX^i_j$ across all clients, where $p_i$ is the proportion of samples trained by client $i$.

When $D_i$'s are different distributions, Lynaponov's CLT generalization shows that the sample mean converges in distribution to the Normal distribution if we have finite moments of order $2 + \delta$ for some $\delta > 0$ \cite{feller}. In particular, if there exists a $\delta > 0$ such that\footnote{This does not apply to distributions with infinite variance such as some power law distributions \cite{mor}.}
\begin{equation}
\frac{1}{s{_N}^{2+\delta}}\sum_{i=1}^N E[|X_j^i-\mu_i|^{2+\delta}] \rightarrow 0 \,\, as \,\,  n \rightarrow \infty   
\end{equation}
where $s_N^2 = \sum_{i=1}^{N} \sigma^2_i$. When the distributions $D_i$'s are different, then the averaged parameters sent to clients in the new round may not generalize to other clients (i.e., clients who were not selected for training in a given round).
Thus, in \textsf{FedPrune}, rather than using the average value of each parameter, we randomly draw from the Normal distribution with parameters $\mu=\sum_{i=1}^{N} p_iX^i_j$ and $\sigma^2$, where the latter is the variance across all clients \emph{within} a single round.
In our evaluation, we show that using this strategy improves accuracy, robustness as well as convergence speed.



\subsection{Algorithm}
\SetKwComment{Comment}{$\triangleright$ }{}

{\SetAlgoNoLine%
\begin{algorithm}
\caption{\textsf{FedPrune}}\label{alg:two}

\KwIn{Model Dropout Rate ($k\%$), Pruning Round ($r$)}
\textbf{Server executes:}\\
\textbf{Initialize:} Global model $W_0$, mask $M \gets 0$\; 
\For{each round t = 1, 2, . . . , T}{
    \eIf{$t$ > 1}{
        Select sub-model $w_t$ from $W_t$ based on mask $M$ \\
    }{
        
        $w_t \gets \text{Random selection } k\%$ \;
        $M \gets \text{Indexes of sub-model }w_t$\;
    }
    $C_t \gets \text{(select $n$ clients randomly)}$ \Comment*[r]{n $\leq$ N} 
    Send $W_t$ or $w_t$ to $C_t$ based on their device characteristics \\
    \For{each client $c \in C_t$}{ 
        \eIf{$c$ is slow}{
            \textbf{Train sub-model $w_t$:} \\
            $activations^c_{t},  w^c_{t+1} = \ell(w_t, c)$ \;
            $W^c_{t+1} = Broadcast(w^c_{t+1}, M)$
            
        }{
            \textbf{Train large model $W_t$:} \\
            $activations^c_{t},  W^c_{t+1} = \ell(W_t, c)$ \;
        }
    }
    $activations_t = \frac{1}{n}\sum_{c \in C_t} activations^c_{t}$\;
    $\mu = \sum_{c \in C_t} \frac{s_c}{S} W^c_{t+1}$ \Comment*[r]{$S_t$: Total samples} 
    $\sigma = \text{WeightedStdev}(W_{t+1})$\;
    $\sigma = \frac{\sigma}{\sqrt{t}}$\;
    $W_{t+1} = \mathcal{N}(\mu,\,\sigma^{2})\,$
    
    \If{$(t \mod r) = 0$}{
        Update $M$ using activations of dense layer and $\ell$1-Norm of CNN filters   
    }
}
\end{algorithm}}%

At start of the learning process, the server initializes a global model $W_0$ and a mask $M$. The server randomly picks a smaller model $w_0$ (which we call a \emph{sub-model}) from the global model $W_0$ to send to slow clients. It does so randomly at the start as it does not have any metric to effectively choose a sub-model. Mask $M$ is used to keep track of the random indices. The server selects $n$ clients randomly and sends the sub-model $w_0$ or the large model $W_0$ to selected clients based on their device characteristics. Let $C_t$ denotes clients in $t^{th}$ round. Each client trains its model and sends the model updates as well as activations of dense layers to the server. The server aggregates these updates and activations. In weighted aggregation, $s_c$ and $S_t$ represent number of training samples of client $c$ and total number of training samples in round $t$ respectively. Based on average activations, and $\ell$1-norm of the CNN filters, the server picks an optimal sub-model for slow clients after every $r$ rounds. The only two parameters of \textsf{FedPrune} are the Model Dropout Rate (MDR) $k$, representing the percentage of neurons and filters to be dropped from dense and convolution layers respectively when constructing the sub-model and Mask Update Round (MUR) $r$, at which round to update model dropout mask.

\textbf{Aggregation:} \textsf{FedPrune} uses insights from the Central Limit Theorem (CLT) to aggregate clients' model parameters. Server calculates the weighted mean $\mu$ (weighted by number of data samples) and standard deviation $\sigma$ of parameters. It then uses this mean $\mu$ and standard deviation $\sigma$ to randomly sample parameters from the normal distribution to send back to the clients. These randomly sampled parameters generalize better even when the distribution is non-IID. 
As the training progresses, the model becomes stable and large changes in model parameters can adversely impact performance. As a result, we continue decreasing $\sigma$ proportional to $1/\sqrt{t}$ (current round), which does not let model deviate significantly from the stable parameters yet helps in better generalization.
\section{Evaluation}
\label{sec:eval}
In this section, we present empirical results for \textsf{FedPrune} using large-scale simulations and small-scale real testbed experiments involving multiple mobile clients under federated settings. 
We demonstrate the effectiveness of \textsf{FedPrune} in the presence of system and statistical heterogeneity and study its convergence and robustness properties.
All code, data, and scipts for generating our results will be made publicly
available on GitHub.
We provide details of our experimental setup in Section 5.1 and Appendix \ref{leaf_datasets_models}.

\subsection{Experimental Details}
We evaluate \textsf{FedPrune} on multiple models, tasks and real-world federated datasets. We use LEAF \cite{leaf}, a benchmarking framework for federated learning to simulate our federated setup. We implement \textsf{FedPrune} in LEAF and evaluate its performance on convolution neural networks, LSTMs models and explore four real world datasets in federated learning setting. Specifically, we use CIFAR-10, Federated extended MNIST (FEMNIST) and Fashion-MNIST (FMNIST) for CNN and Sent140 for LSTM models.

\begin{table}[t]
\caption{Statistics of four real federated datasets}
\centering
\begin{tabular}{c c c c c}
\hline\hline
Dataset & Devices & Samples & Mean & Stdev \\ [0.5ex] 
\hline
FEMNIST & 206 & 30774 & 149 & 59 \\
FMNIST & 500 & 72505 & 145 & 138 \\
CIFAR-10 & 250 & 60000 & 240 & 0 \\
Sent140 & 772 & 40,783 & 53 & 32 \\ [1ex]
\hline
\end{tabular}
\label{table:dataset_detail}
\vspace{-0.1in}
\end{table}
\textbf{Real Data.}
We use four real-world datasets; whose statistics are summarized in Table \ref{table:dataset_detail}. These datasets are curated from prior work in FL \cite{fedprox,ditto,flclient} as well as recent FL benchmarks in LEAF \cite{leaf}. FMNIST, FEMNIST, and Sent140 are non-IID datasets. To study the behaviour of \textsf{FedPrune} under a IID dataset, we curated CIFAR-10 in a IID fashion, where each examples has the same probability to belong to any device.  We then study a more complex 62-class FEMNIST dataset \cite{cohen,flclient}.
To study statistical heterogeneity and its impact, we experiment with two versions of the FEMNIST dataset. One version is used in a prior work \cite{ditto} whereas the second version is skewed NIID in which we randomly assign 5 classes to each device. For the non-convex setting, we consider a text sentiment analysis task on tweets from Sentiment140 (Sent140) with an LSTM classifier, where each twitter account corresponds to a device \cite{sent140}.
Details of datasets, models, and workloads are provided in Appendix \ref{leaf_datasets_models}.

\textbf{Implementation.}
As LEAF does not provide functionality for simulating individual slow and fast clients, we update LEAF FedAvg implementation to include slow clients that are unable to complete model training in time and thus get dropped. In order to draw a fair comparison with FedAvg, we employ SGD as the local solver for \textsf{FedPrune} and average the parameters after every round with weights proportional to the number of local data points.

\textbf{Hyperparameters.}
For each dataset, we use learning rates from \{0.01, 0.001, 0.0003\}. 
We observe that the behavior of all techniques remains the same relative to each other but a smaller learning rate means the model takes longer to converge. We report results with 0.001 learning rate for all experiments on CNN datasets. For Sent140, previous works have picked a large learning rate, therefore, we also use 0.01 learning rate for Sent140. We set the number of selected devices in one round to be 10. Batch size is also 10 for all experiments. We set the \textsf{FedPrune} parameters $k=50\%$ and $r=10$ for our experiments. In case of CNN, 50\% filters and neurons are dropped from convolution and dense layers respectively. While the pruned model of LSTM has 50\% less cells in the LSTM layers for slow client. For fair comparison, we fix the randomly selected devices, the slow clients, and mini-batch orders across all runs. We take 10 runs for each experiment and report average results.

\begin{figure}[t]
    \centering
    \includegraphics[width=0.75\linewidth,scale=0.35]{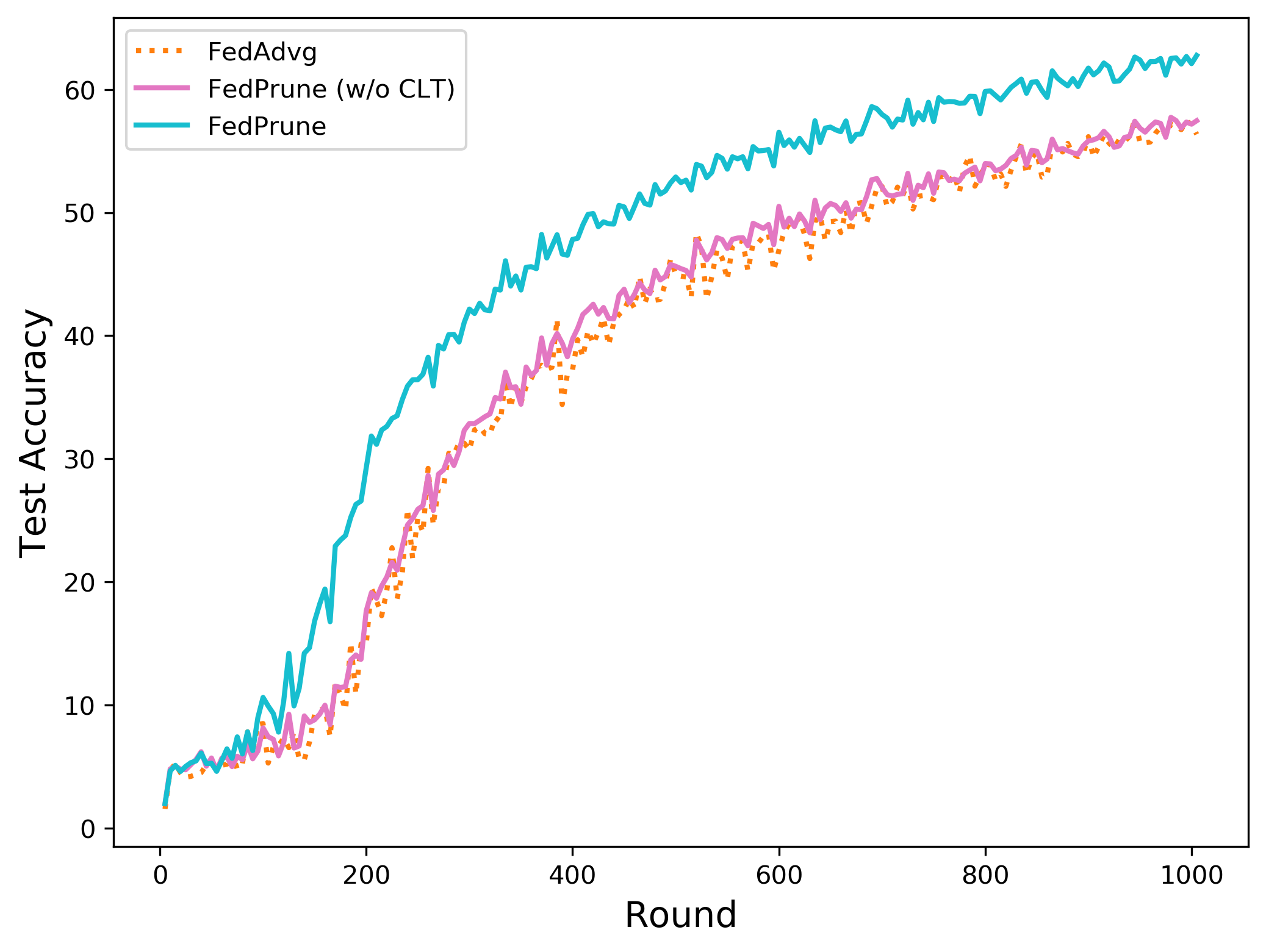}
    \vspace{-0.1in}
    \caption{FEMNIST (skewed): 50\% Client Dropout.}
    \label{fig:feddrop_cnn_lr001_eval05_c10_d05_md05_ep10_b10_femnist_niid}
    \vspace{-0.1in}
\end{figure}
\begin{figure}[t]
    \centering
    \includegraphics[width=0.75\linewidth,scale=0.35]{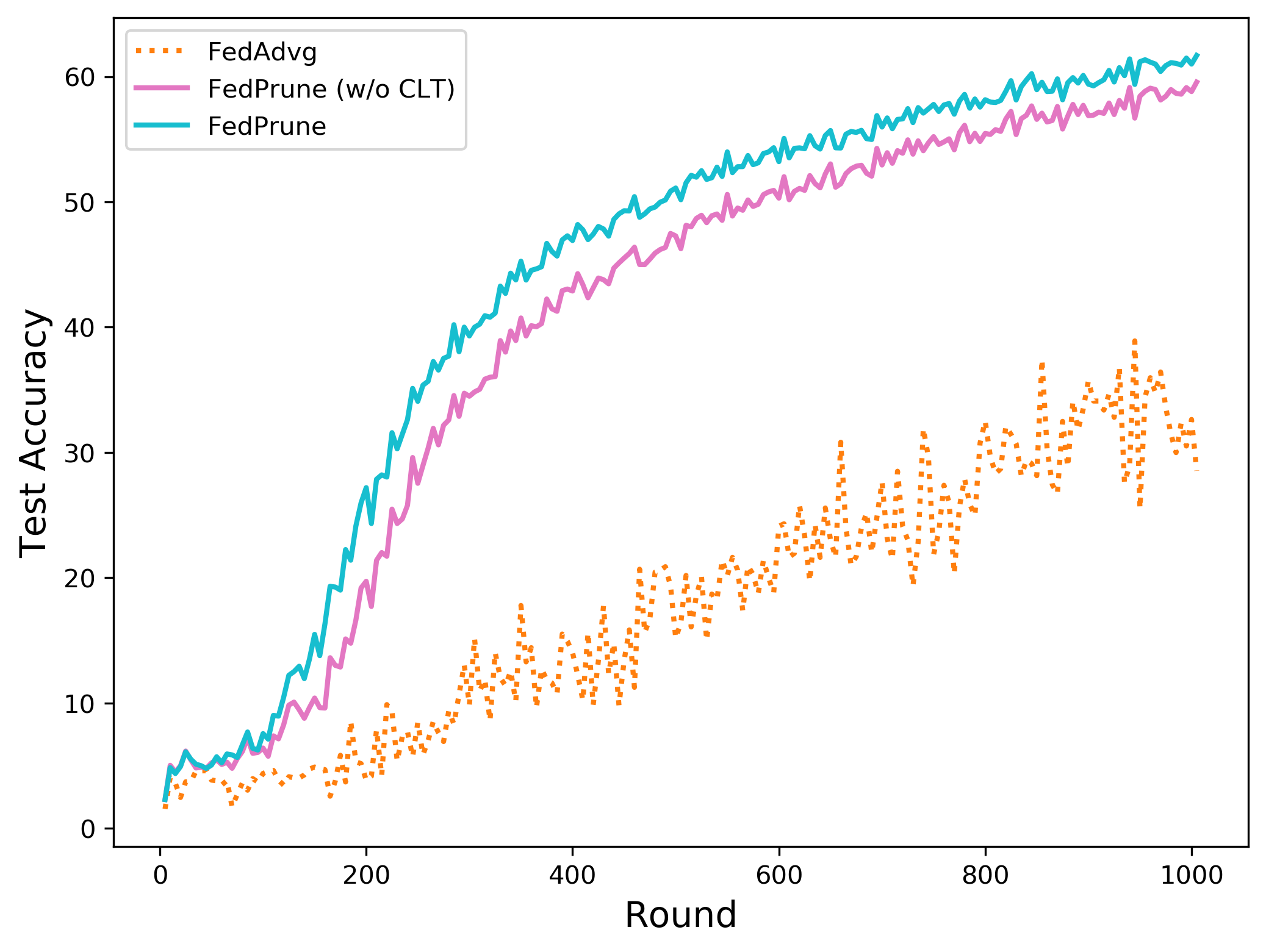}
    \vspace{-0.1in}
    \caption{FEMNIST (skewed): 90\% Client Dropout.}
    \label{fig:feddrop_cnn_lr001_eval05_c10_d09_md05_ep10_b10_femnist_niid}
    \vspace{-0.1in}
\end{figure}

\subsection{System \& Statistical Heterogeneity}
In order to study the impact of system and statistical heterogeneity, we experiment with varying system heterogeneity and diverse datasets.

\textbf{System heterogeneity.}
We assume that there exists a global clock cycle that keeps track of when to start the next round. However, devices participating in a round are limited by their system constraints, which determine the amount of work a device can do in a given time. In our simulations, we fix a global number of epochs $E$ and emulate some devices as slow which could not train $E$ epochs due to their system constraints. These devices get dropped in FedAvg but in \textsf{FedPrune} we send a smaller model to these devices so they can train it in a resource-constrained environment within a given time. For varying heterogeneous settings, we 
experiment with 0\%, 50\%, and 90\% of resource-constrained (slow) devices. Settings where 0\% devices are slow correspond to the environments without systems heterogeneity, while 90\% of the slow devices correspond to highly heterogeneous environments. FedAvg will simply drop these 0\%, 50\%, and 90\% slow clients upon reaching the global clock cycle, and \textsf{FedPrune} will incorporate the updates from these devices as these devices train a subset model and will be able to send updates on time. 

\textbf{Statistical heterogeneity.}
To evaluate \textsf{FedPrune} under statistical heterogeneity, we use datasets with varying degree of IID-ness. 
Our results show that model accuracy and convergence becomes worse as statistical heterogeneity increases (see Figures \ref{fig:feddrop_cnn_lr001_eval05_c10_d05_md05_ep10_b10_femnist_niid} and \ref{fig:feddrop_cnn_lr001_eval05_c10_d09_md05_ep10_b10_femnist_niid}).
However, \textsf{FedPrune} can better handle statistical heterogeneity as it employs a CLT based approach to aggregate parameters from clients, which helps in improving robustness.
While FedAvg aggregates parameters by weighted averaging, \textsf{FedPrune} uses a different approach. It first computes the mean and standard deviation (stdev) of clients' parameters and then randomly samples from the Normal distribution with that mean and stdev. 
As these parameters are from different distributions, aggregation of these parameters will converge to Normal according to CLT. 
Figures \ref{fig:feddrop_cnn_lr001_eval05_c10_d05_md05_ep10_b10_femnist_niid} and \ref{fig:feddrop_cnn_lr001_eval05_c10_d09_md05_ep10_b10_femnist_niid} show that \textsf{FedPrune} achieves improved performance by using this approach. As the degree of NIID-ness increases, we see that \textsf{FedPrune} with CLT achieves better generalization. In case of high system and statistical heterogeneity (Figure \ref{fig:feddrop_cnn_lr001_eval05_c10_d09_md05_ep10_b10_femnist_niid}), \textsf{FedPrune} provides 22.7\% test accuracy improvement over FedAvg.
\begin{figure*}[t]
  \includegraphics[width=1\linewidth,scale=0.5]{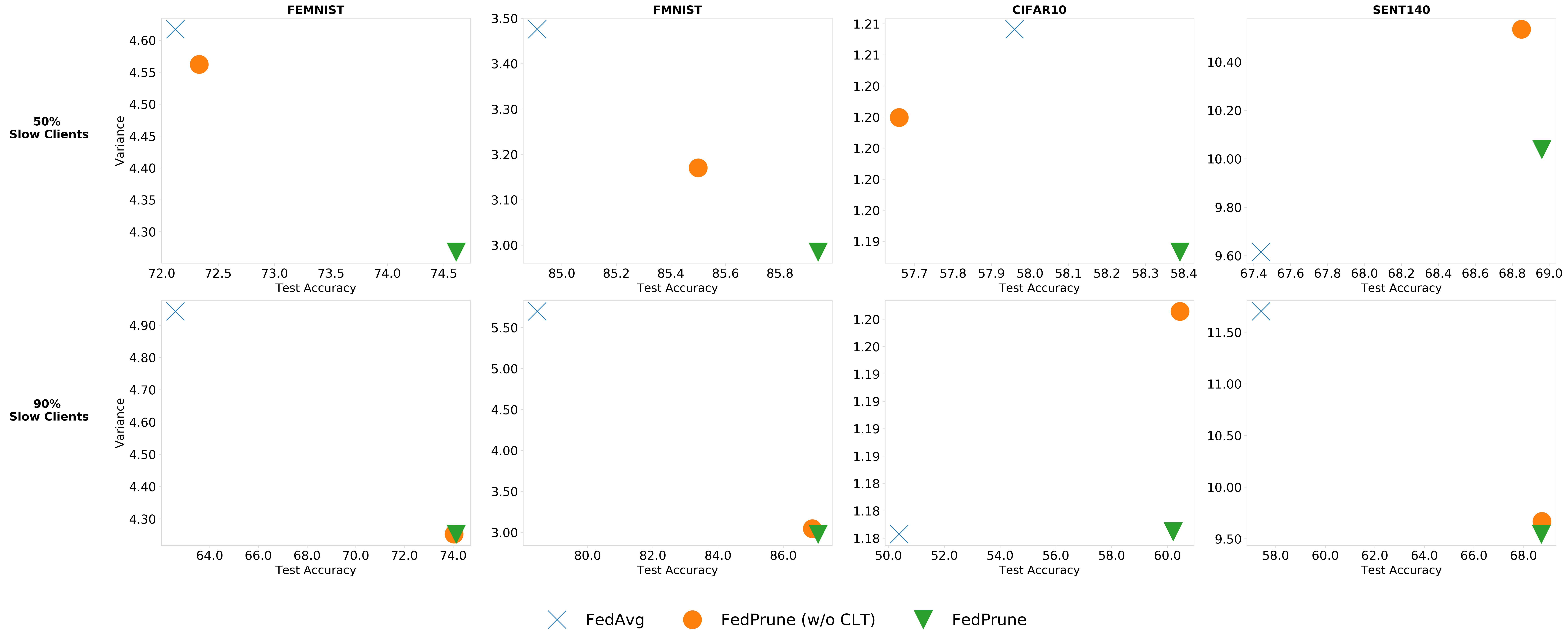}
  \caption{Variance vs Test Accuracy.}
  \label{fig:variance_vs_test_accuracy}
  \vspace{-0.1in}
\end{figure*}

\textbf{Fairness.}
Due to statistical heterogeneity in federated networks, the performance of a model may vary significantly across different devices, resulting in \emph{representation disparity} \cite{motto}.
In \textsf{FedPrune}, we serve a subset model to slow clients, which potentially has a larger risk of representation disparity. We empirically show that in addition to improving accuracy, \textsf{FedPrune} also offers improved fairness. \textsf{FedPrune} picks the best subset model after every $r$ ($r$ is a tuneable parameter) rounds for slow clients.
Variance of test accuracy across clients can be seen in Figure \ref{fig:variance_all}, Appendix \ref{simulated_experiments}. Interestingly, \textsf{FedPrune} provides better average test accuracy as well as achieves lower variance across clients compared to FedAvg and \textsf{FedPrune} without CLT.
Figure \ref{fig:variance_vs_test_accuracy} compares the robustness and fairness of \textsf{FedPrune}, \textsf{FedPrune} w/o CLT and FedAvg. Results show that \textsf{FedPrune} provides robust and fair performance as it trains on all clients and uses CLT to better generalize on data.

\subsection{Real Testbed Experiments}
\label{leaf_model_architecture}
We implement \textsf{FedPrune} on a real FL testbed.
We use PySyft \cite{pysyft}, an open-source framework for FL, to write models and scripts using PyTorch that get executed on mobile devices. PySyft models are hosted on PyGrid \cite{pygrid}, a
flask-based server that in addition to hosting models, performs averaging of weights learned from all clients. Mobile devices connect with the server to download models and train them on the device using KotlinSyft \cite{kotlinsyft}.
The server communicates with the clients using Google Firebase Services \cite{firebaseservices}. We used Android Studio for customizations in KotlinSyft. We perform evaluations on actual mobile devices i.e., Nexus 6P (3\,GB RAM, Octa-core) and Nokia 1 (1\,GB RAM, Quad-core) as fast and slow client, respectively. 
We use real datasets i.e., FEMNIST and a CNN model from the LEAF benchmark to evaluate the impact of model size on model training times. The model has 6.6 million parameters that includes two \textit{CNN} layers and two \textit{dense} layers. The number of \textit{filters} in the first and second \textit{CNN} layers are 32 and 64, respectively with same filter size i.e., (5,5). The number of \textit{neurons} in the first and second \textit{dense} layers are 3136 and 2048, respectively.

Figure \ref{fig:training_time_by_pruning} shows the training time against different model drop rates on the slow device. Large model refers to the complete model being served to the slow client, 30\% drop rate refers to the case when the model is 30\% pruned, and so on. We observe that the training time decreases as we increase the model drop rate. With a drop rate of 50\%, training time is reduced by 66.7\%. This reduction in the training time is due to reduced FLOPs as shown in Figure \ref{fig:flops_by_pruning}.
In particular, a model drop rate of 50\% reduces FLOPs by 3.8$\times$, which in turn improves training time by $\sim$2.9$\times$.

\section{Discussion}
\textbf{Adaptive model serving}.
We evaluate \textsf{FedPrune} using a 2-model approach (i.e., fast clients train over the global model whereas slow clients train over a sub-model). In the future, it would be useful to examine the effectiveness of customizing model sizes for \emph{each} client based on their characteristics.

\noindent
\textbf{\textsf{FedPrune and multi-task learning}}. 
In multi-task learning \cite{multitask}, the goal is to train \emph{personalized} models for each device independent of sizes whereas \textsf{FedPrune} focuses on reducing the overhead of model training for improving inclusiveness in the presence of client heterogeneity.

\noindent
\textbf{Synchronous vs. asynchronous FL}. While we focus on synchronous FL in this work, asynchronous FL is an alternative approach for mitigating stragglers. We do not consider the latter approach in part because the bounded-delay assumptions associated with most asynchronous schemes limit fault tolerance \cite{asyncFL,multitask}.
However, it would be interesting to explore the relative merits of asynchronous methods and synchronous FL in future work.


\section{Conclusion}
We presented the design and evaluation of  \textsf{FedPrune}, an inclusive framework for federated learning that achieves improved learning and fairness properties in the presence of client device heterogeneity.
\textsf{FedPrune}'s adaptive model serving ensures that slow clients are not dropped from the training process and achieve training times similar to fast clients whenever possible. 
Our evaluation involving large-scale simulations and a small-scale real testbed of mobile clients shows that \textsf{FedPrune} achieves robust performance across a variety of real-world federated datasets.

\bibliographystyle{abbrv}
\bibliography{paper,main}

\appendix
\clearpage
\section{Appendix}
\subsection{Datasets and Models}
\label{leaf_datasets_models}
Here we provide complete details of datasets used in our experiments. We experiment with diverse set of real world datasets.

\textbf{FEMNIST.}
This is federated version of extended MNIST dataset which consists of images of 28 x 28. 
FEMNIST is an image classification dataset with 62 classes. This data is partitioned by writer of the digits/characters. We train a CNN model on this dataset. 
We have two versions of femnist dataset with different levels of statistical heterogeneity. One version is less heterogeneous and used by Ditto \cite{ditto}. While we create the other version in LEAF by assigning 5 random classes to each client. We show that the \textsf{FedPrune}
can be more beneficial in the face of statistical heterogeneity i.e., skewed FEMNIST data (figure \ref{fig:feddrop_cnn_lr001_eval05_c10_d05_md05_ep10_b10_femnist_niid} \and \ref{fig:feddrop_cnn_lr001_eval05_c10_d09_md05_ep10_b10_femnist_niid}). For FEMNIST, we use a model with two convolutional layers which are followed by pooling layers, and a final dense layer.

\textbf{FMNIST.}
We experiment on Fashion MNIST dataset used by Ditto \cite{ditto}. It is also an image classification dataset which has 10 classes. We use the same model for FMNIST that we described above for FEMNIST.

\textbf{CIFAR-10.}
CIFAR-10 is also an image classification dataset. To create a federated version of CIFAR-10, we randomly assign an image to a client. Each image has the same probability to belong to any client, which makes this dataset IID. CIFAR-10 model architecture is also same as FEMNIST.

\textbf{SENT140.}
Sentiment140 \cite{sent140} is a text dataset of tweets which is used for sentiment analysis task. We perform our experiments on sent140 used in FedProx \cite{fedprox}. We train a model with two LSTM layers and a fully connected layer. We use a pretrained 300D GloVe embedding \cite{glove}. Glove embeds each character  of the input sequence into a 300-dimensional space
and outputs one character per training sample after 2 LSTM layers and a fully-connected layer.

\subsection{Real Device Experiments}
\textbf{Quantifying training time against different model sizes}
\label{synthetic_results}
We performed the experiments for different model sizes on synthetic and femnist dataset to quantify training time on actual devices.

\begin{figure}[t]
    \centering
    \includegraphics[width=0.9\linewidth]{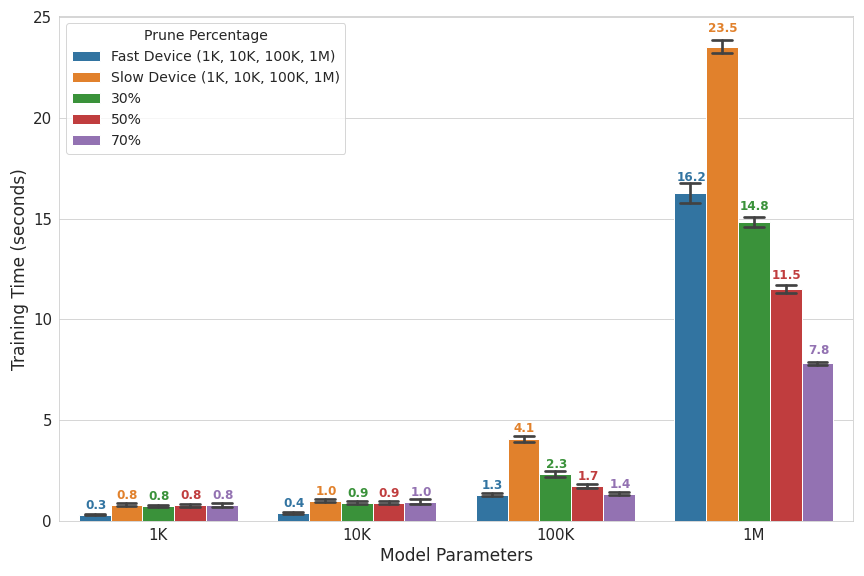}
    \caption{Training time against different model sizes for synthetic dataset}
    \label{fig:training_time_by_model_sizes}
\end{figure}

\begin{figure}[t]
    \centering
    \includegraphics[width=0.9\linewidth,scale=0.35]{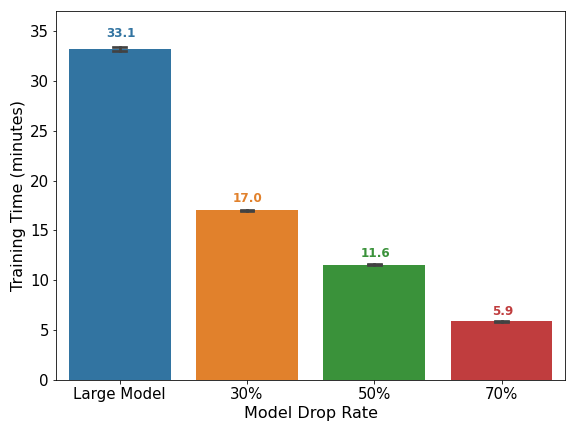}
    \vspace{-0.1in}
    \caption{Training times under \textsf{FedPrune} with different model sizes.}
    \label{fig:training_time_by_pruning}
   \vspace{-0.1in}
\end{figure}

Figure  \ref{fig:training_time_by_model_sizes} shows the training time of linear model on slow and fast devices against different model sizes. Blue bar represents the training times of fast device, orange represents when a large model was served to the slow client, green bar represents when 30\% pruned was served, and so on. Slow client takes $\sim$1.4x - 3.1x more time as compared to fast client. We observe that the model training time is reduced by 1.4x compared with fast client if we serve the 50\% reduced model to the slow client. Figure \ref{fig:training_time_by_pruning} shows the training time of CNN model on femnist dataset for different model drop rates.

\begin{figure}[t]
    \centering
    \includegraphics[width=0.9\linewidth,scale=0.35]{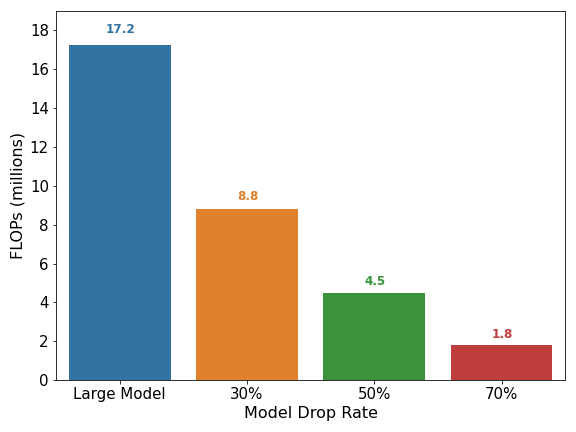}
       \vspace{-0.1in}
    \caption{FLOPS for different model drop rates.}
    \label{fig:flops_by_pruning}
      \vspace{-0.1in}
\end{figure}

\subsection{Simulation Experiments}

\label{simulated_experiments}
Following plots show the training loss and variance of test accuracy across clients on multiple real datasets under various levels of system heterogeneity.
\begin{figure*}[h]
  \includegraphics[width=1\linewidth,scale=0.5]{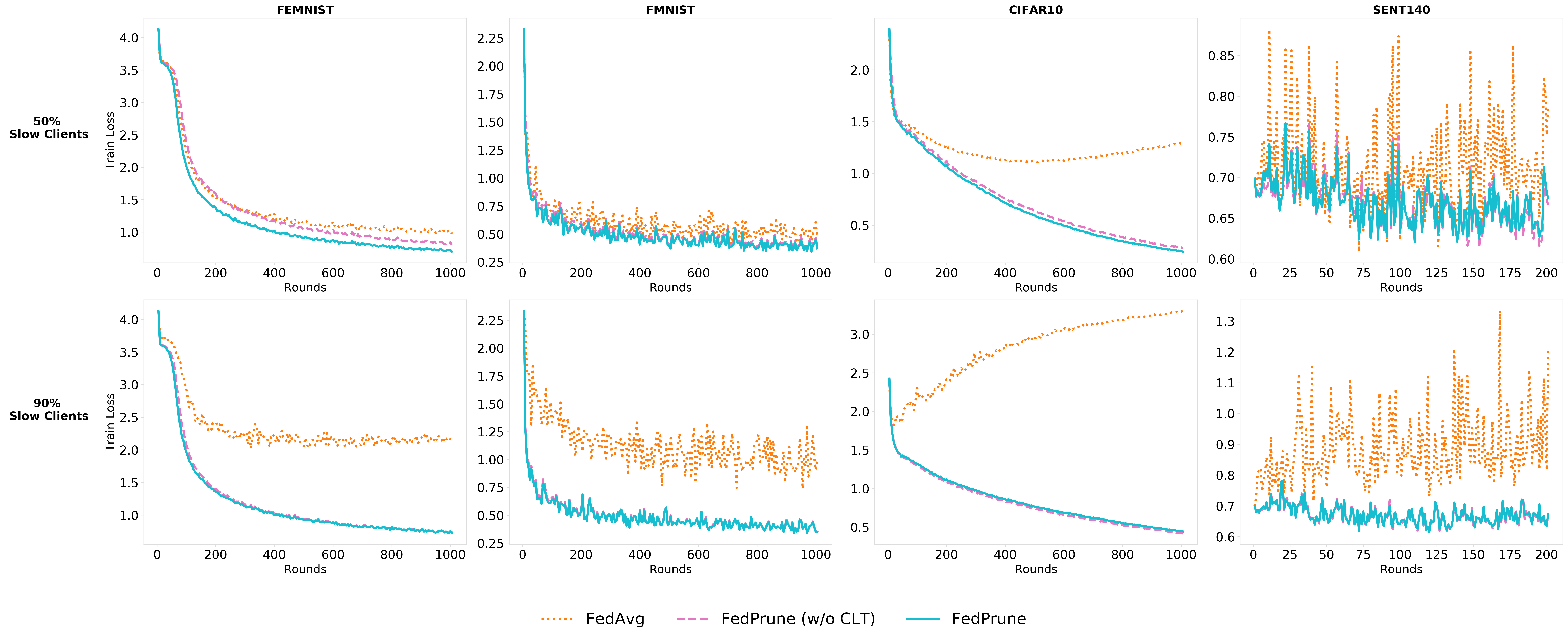}
  \caption{Train loss of various datasets and different levels of systems heterogeneity indicates that \textsf{FedPrune} results in significant convergence improvements relative to FedAvg.}
  \label{fig:train_loss_all}
\end{figure*}

\begin{figure*}[th]
  \includegraphics[width=1\linewidth,scale=0.5]{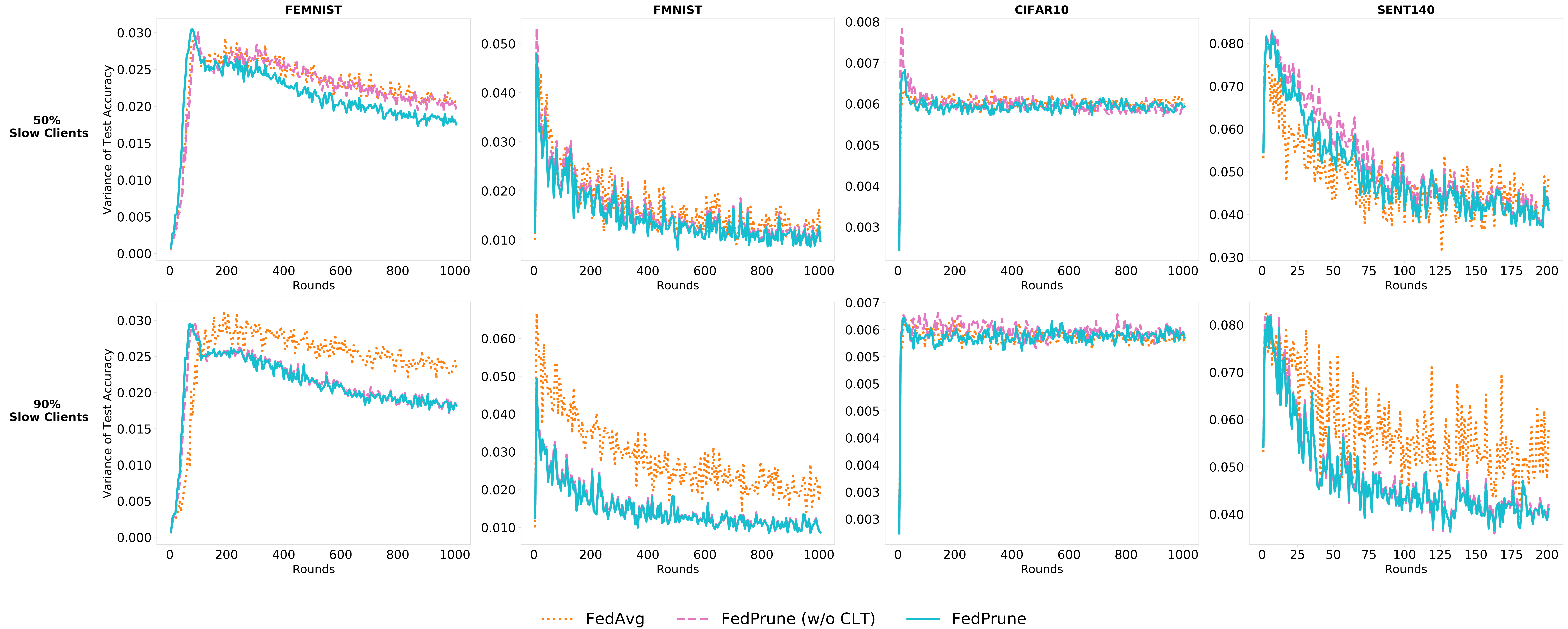}
  \caption{Variance of average test accuracy across clients}
  \label{fig:variance_all}
\end{figure*}

\end{document}